\ifdefined\XeTeXversion
\else
  \pdfoutput=1
\fi
\documentclass[11pt]{article}

\usepackage[final]{acl}

\usepackage{times}
\usepackage{latexsym}
\usepackage[T1]{fontenc}
\usepackage[utf8]{inputenc}
\usepackage{microtype}
\usepackage{graphicx}
\usepackage{booktabs}
\usepackage{amsmath}
\usepackage{amssymb}
\usepackage{multirow}
\usepackage{xcolor}
\usepackage{url}

\title{ComplexConstraints and Beyond: Expert Rubrics for RLVR}

\author{Sushant Mehta\thanks{Correspondence to: \texttt{sushantmehta@surgehq.ai}}, Liudas Panavas, Suhaas Garre, Edwin Chen \\
Surge AI}

\begin{document}
\maketitle
\begin{abstract}
Evaluation protocols can lag behind LLM capabilities. Programmatically verified benchmarks cover narrow surface constraints, whereas real-world instruction following and agentic workflows require judging semantic, contextual, and policy-dependent behavior. We study expert-curated rubric-based evaluation as a unified mechanism for measurement and reinforcement-learning rewards across two settings: complex instruction following and enterprise agentic tasks. We identify rubric-design choices that affect reward quality, including \textit{maximum viable atomicity}, \textit{intent-aware criterion design}, and \textit{LLM-judge calibration}. We introduce \textsc{ComplexConstraints}, an expert-curated instruction-following suite comprising a public 75-prompt benchmark with 1,559 rubric criteria and a disjoint 1,000-prompt training set, with 10--40 atomic criteria per prompt. Empirically, rubric rewards improve training in both fixed task datasets, such as \textsc{ComplexConstraints}, and stateful RL environments, such as CoreCraft. Training a 4B model on \textsc{ComplexConstraints} improves mean criterion pass rate by +15.5 pp on a held-out split, bringing it within 0.5 pp of the untrained baseline of a roughly 60$\times$ larger Qwen3 model, and the gains transfer to external benchmarks the model never saw during training: +8.4 pp on AdvancedIF and +10.1 pp on MultiChallenge. In CoreCraft, rubric-reward RL likewise transfers to out-of-distribution benchmarks (+4.5 pp BFCL, +7.4 pp $\tau^2$-Bench, +6.8 pp Toolathlon). These results show that expert-authored rubrics provide effective evaluation targets and scalable reward signals for improving LLM instruction following and agentic behavior.
\end{abstract}

\section{Introduction}

Traditional benchmarks are increasingly saturated, with frontier models topping out on tests where meaningful capability differences can no longer be distinguished \citep{eriksson2025trustbenchmarks}. Data contamination can further erode confidence in reported scores. More fundamentally, there is a growing misalignment between what benchmarks measure and what real-world deployment requires: namely, the ability to follow complex, layered instructions; to maintain behavioral constraints across extended interactions; and to execute multi-step professional workflows reliably.

IFEval \citep{zhou2023ifeval}, one of the most widely cited instruction-following benchmarks, illustrates this gap vividly. Its 25 verifiable constraint types (word counts, forbidden characters, formatting rules) can be checked programmatically, but this programmatic verifiability comes at the cost of validity. A model can produce incoherent nonsense and still pass, so long as it avoids commas and the letter ``c.'' The benchmark shaped itself around the evaluation method rather than around the construct it claims to measure.

This paper examines an alternative paradigm: \textit{expert-curated rubric-based evaluation}, in which domain experts decompose task success into atomic, verifiable criteria that capture both explicit requirements and pragmatic user intent. We analyze this approach across two complementary domains: \textbf{complex instruction following}, using \textsc{ComplexConstraints}, a new expert-curated benchmark and training suite we introduce in this work, alongside the independently developed AdvancedIF and MultiChallenge benchmarks \citep{he2025advancedif, deshpande2025multichallenge}; and \textbf{agentic task execution}, using the CoreCraft enterprise simulation \citep{mehta2026corecraft, ritchie2026hierarchy}.

Our paper makes three contributions. First, we describe rubric-design decisions that improve construct validity relative to programmatic verification and make expert rubrics useful as reward signals (\S\ref{sec:principles}). Second, we introduce \textsc{ComplexConstraints}, an expert-curated instruction-following suite embodying these decisions: a public 75-prompt benchmark with 1,559 expert-authored rubric criteria, and a disjoint 1,000-prompt training set built through the same pipeline, with 10--40 atomic criteria per prompt. Third, we present empirical evidence demonstrating that expert rubrics designed according to these decisions are highly effective as reward signals for reinforcement learning (\S\ref{sec:training}): in both instruction following and agentic task execution, rubric-based RL training on modest amounts of data produces substantial, transferable improvements, suggesting that expert rubrics serve not only as more valid evaluation instruments but also as highly effective training signals.

\section{Background and Related Work}
\label{sec:background}

\paragraph{From Programmatic to Rubric-Based Evaluation.}
Instruction-following evaluation has progressed through several stages. IFEval \citep{zhou2023ifeval} established programmatic verification with 25 constraint types. FollowBench \citep{jiang2024followbench} revealed that model performance degrades as constraint complexity increases. InFoBench \citep{qin2024infobench} proposed decomposing complex instructions into binary verification questions. ComplexBench \citep{wen2024complexbench} introduced a hierarchical taxonomy of constraint composition. MultiChallenge \citep{deshpande2025multichallenge} extended realistic evaluation to multi-turn conversation, grading with instance-level rubrics. AdvancedIF \citep{he2025advancedif} moved to expert-curated rubrics for 1,600+ prompts, finding that even frontier models achieve only approximately 75\% when assessed against these richer criteria. In a complementary medical setting, HealthBench \citep{arora2025healthbench} pairs 5,000 multi-turn health conversations between models and users with 48{,}562 physician-written rubric criteria across multiple behavioral axes (accuracy, communication, context awareness). Our \textsc{ComplexConstraints} dataset, introduced in this paper, shares the expert-authored rubric design of these recent efforts but is built from the start to serve a second purpose: the same rubrics that evaluate frontier models also function as data-efficient reward signals for reinforcement learning. The broader progression reflects a general trend: as evaluation criteria become more expressive, they better distinguish model capabilities, but they also become harder to automate reliably. Our results suggest that this investment in expressiveness pays a second dividend: rubrics rich enough to evaluate frontier models are also rich enough to train them.

\paragraph{LLM-as-a-Judge.}
Rubric-based evaluation typically relies on LLM judges to assess criterion satisfaction. A growing body of work examines the reliability of this approach. \citet{gu2024llmjudgesurvey} provide a comprehensive survey, while \citet{yamauchi2025empiricalstudy} find that clear evaluation criteria matter more for judge reliability than chain-of-thought prompting, which adds little once such criteria are present. \citet{schroeder2024reliability} demonstrate limitations of single-shot LLM evaluations, and \citet{chehbouni2025neithervalid} argue from a measurement-theoretic perspective that LLM-judge validity remains undertested. \citet{rao2026autorubric} consolidate many of these design choices into Autorubric, a unified open-source framework for rubric-based LLM evaluation that operationalizes per-criterion atomic evaluation, ensemble judging, and psychometric reliability metrics. These concerns motivate careful rubric design that reduces evaluator ambiguity.

\paragraph{Agentic Evaluation.}
Agent benchmarks have evolved from simplified web interfaces toward realistic, execution-based environments \citep{zhou2024webarena}. Function calling is tracked by the Berkeley Function Calling Leaderboard (BFCL) \citep{patil2025bfcl}, and customer service evaluation is addressed by $\tau$-bench \citep{yao2024taubench} and its successor $\tau^2$-Bench \citep{barres2025tau2bench}. Toolathlon \citep{li2025toolathlon} benchmarks agents on 108 diverse, long-horizon tasks. A survey of 86 practitioners working with deployed agent systems found that 68\% of deployed agents execute ten or fewer steps before human intervention \citep{pan2025measuring}, underscoring the gap between benchmark performance and deployment readiness.

\paragraph{Rubrics as Training Signals.}
Reinforcement Learning from Verifiable Rewards (RLVR), introduced by T\"ulu~3 \citep{lambert2024tulu3} and popularized by DeepSeek-R1 \citep{guo2025deepseekr1}, has been extended to instruction following through rubric-based reward signals. RIFL \citep{he2025advancedif} uses a finetuned rubric verifier to provide rewards for RL training. VerIF \citep{peng2025verif} combines rule-based and LLM-based verification. RLCF \citep{viswanathan2025rlcf} extracts instruction-specific checklists. ToolRL \citep{qian2025toolrl} demonstrates that GRPO-based training with fine-grained tool-use rewards enables tool-use generalization.

\paragraph{Concurrent Rubric-Based Work.}
Several recent and concurrent efforts use rubrics for either evaluation or RL training, and \textsc{ComplexConstraints} occupies a distinctive position in this landscape along two axes: \emph{rubric authorship} (human-expert vs.\ LLM-generated) and \emph{purpose} (evaluation only vs.\ RL training signal). RubricRAG \citep{dhole2026rubricrag} addresses the cost of human authoring by retrieving domain knowledge to generate query-specific rubrics at inference time, RubricBench \citep{zhang2026rubricbench} quantifies the quality gap between model-generated and human rubrics, and Autorubric \citep{rao2026autorubric} unifies the judging infrastructure. Relative to these, \textsc{ComplexConstraints} contributes (i) \emph{expert-authored} rubrics, deliberately chosen over synthetic generation to capture pragmatic intent that LLM-generated rubrics tend to miss; (ii) coverage of \emph{instruction following with high constraint density} (10--40 criteria per prompt), complementary to the medical focus of HealthBench and the evaluation-only orientation of AdvancedIF's benchmark component; and (iii) explicit \emph{dual-purpose} design: the same rubrics serve as evaluation instruments and as RL reward signals. The design discussion in \S\ref{sec:principles} consolidates lessons from this line of work into guidance for expert rubric authoring.

\section{What Makes a Rubric Trainable}
\label{sec:principles}

Naive implementations of using a rubric as a reward can fail in several predictable ways. Partial credit can flow to confidently wrong answers. Surface-anchored criteria can reward phrasing over intent. Ambiguously phrased criteria can turn the judge into a source of noise rather than signal.

Our rubric methodology exists to close these failure modes. We highlight the design decisions that most directly determine reward quality when the rubric is used as a reward signal during RL.

\subsection{Maximum Viable Atomicity}

Standard rubric guidance holds that criteria should be atomic: each should test one thing. Applied literally, atomicity degrades the signal. Our criteria instead target the smallest meaningful unit. For evaluation, this keeps scores correlated with response quality. For RL, the distinction determines whether the gradient points toward correct answers or toward responses that are confidently wrong in superficially similar ways.

\subsection{Intent over Literalism}

A user asks for help improving their Spanish, while fluently summarizing an economics article they are reading in Spanish. A rubric that takes ``improve my Spanish'' at face value rewards beginner vocabulary lists. Used as a reward, the failure compounds beyond a single misgraded response: the rubric systematically trains the model to privilege surface phrasing over context. Before writing any criteria, our rubric authors therefore read several model responses to the prompt and restate the user's intent in their own words -- the ``what'' and the ``why'' -- then write criteria against that restatement. Intent-aware criteria make contextual reading a rewarded behavior.

\subsection{Calibrating the Judge}

Every criterion is validated against an LLM verifier before deployment, through an iterative loop. The author drafts the criterion, grades a reference response by hand, and checks that the verifier agrees. On disagreement, the author diagnoses whether the criterion's phrasing is genuinely ambiguous and revises it. The adversarial step follows: the author edits the response so that the correct verdict flips, and confirms that the verifier's verdict flips with it.

This loop can surface ambiguities no style guide anticipates. A criterion asking a poem to ``avoid alliteration'' failed when the verifier flagged ``his heart.'' The fix -- specifying poetic alliteration, the repetition of initial stressed consonant sounds -- produced agreement in both directions. Other rules accumulate the same way: embed answer keys directly in factual criteria so the judge verifies instead of recalling; decompose criteria too broad to grade reliably into concrete, checkable parts.

This step matters even more for training. A noisy judge produces a noisy reward, and RL optimizers locate and exploit reward noise efficiently. Criterion calibration functions as a defense against reward hacking.

\subsection{Optimal Difficulty for RL}

RL extracts the most training signal where models sometimes succeed: tasks that every model passes yield no gradient, and tasks that no model passes yield none either. Because rubric authors evaluate frontier responses against their rubrics during construction, difficulty is tuned task by task into the band where the signal is most informative.

Per-criterion density provides a second mechanism. Passing a task outright requires satisfying every criterion, so full success remains rare even for frontier models. Per-criterion scoring, however, creates a continuous gradient between partial and full success: a response satisfying 28 of 30 criteria receives meaningfully more reward than one satisfying 15, and the optimizer learns from the difference.

\section{The \textsc{ComplexConstraints} Dataset}
\label{sec:complexif}

To validate the design decisions above and to provide a concrete instantiation of expert rubric-based evaluation, we introduce \textsc{ComplexConstraints}, an expert-curated instruction-following suite designed to serve as both a challenging evaluation benchmark for frontier models and an effective training resource for RLVR.

\textsc{ComplexConstraints} comprises two disjoint sets built through the same expert pipeline: a public \emph{benchmark} of 75 prompts with 1{,}559 rubric criteria, used for fine-grained measurement of frontier models,\footnote{Released under CC-BY-4.0 at \url{https://huggingface.co/datasets/surgeai/ComplexConstraints}.} and a companion \emph{training set} of 1,000 single-turn prompts used for the RL experiments in \S\ref{sec:training}. Each prompt is paired with 10 to 40 atomic rubric criteria (median 19) authored by domain experts following the design decisions described in \S\ref{sec:principles}, with each criterion labeled along two axes: explicit vs.\ implicit, and objective vs.\ subjective. Every prompt mirrors authentic professional use cases with high instruction density: constraints that depend on one another (multi-step), fire only under certain conditions (conditional), require thinking ahead (planning), or go unstated (implicit). Tasks are grounded in real-world scenarios such as scheduling under combinatorial constraints, iterative document editing, and multi-requirement content generation; the training set spans seven task categories weighted toward professional writing and planning work: business writing (39\%), scheduling (22\%), data categorization (11\%), personal plans (10\%), numeric processing (7\%), creative writing (6\%), and extraction (4\%).

Both sets consist exclusively of single-turn prompts, each authored to exercise a high density of layered constraints within a single user instruction. Rubric criteria undergo LLM-judge calibration before use as evaluation criteria or reward signals.

\paragraph{Data construction.}
Prompts were authored from scratch by a workforce of domain experts spanning several task categories. Each prompt passed through a three-stage workflow: (i) draft authoring by a domain expert, including a reference response used to anchor rubric writing; (ii) rubric authoring against the reference; and (iii) independent review by a second expert, with adjudication by a senior reviewer when authors disagreed. Difficulty was calibrated through a pilot evaluation against a frontier model.

\paragraph{Difficulty profile.}
Even the strongest frontier model on the public leaderboard passes barely two-fifths of benchmark tasks outright (where passing requires \emph{all} rubric criteria to be satisfied; Table~\ref{tab:complexif}), and most models score far lower. This difficulty profile is not artificially engineered but arises naturally from the realistic complexity of the prompts, making it well-suited for both evaluation and RL training. The continuous gradient between partial and full success, provided by the high rubric density, produces a rich reward landscape for reinforcement learning.

\section{Expert Rubrics as RL Training Signals: Empirical Evidence}
\label{sec:training}

Having established what high-quality rubrics look like, we now turn to the central empirical question: \textit{are expert-curated rubrics effective as reward signals for reinforcement learning?} We present evidence across both instruction following and agentic task execution, showing that rubric-based RL training produces substantial, transferable improvements from modest amounts of data.

\subsection{Targeting Frontier Capability Gaps for Training}
\label{sec:gaps}

\begin{table}[t]
\centering
\small
\begin{tabular}{lc}
\toprule
\textbf{Model} & \textbf{Task Pass \%} \\
\midrule
Gemini 3.1 Pro & 40.4 \\
GPT-5.5 & 38.7 \\
Gemini 3.5 Flash & 36.9 \\
Qwen3.7 Max & 36.0 \\
Claude Opus 4.8 & 34.9 \\
Kimi K2.6 & 34.0 \\
Claude Opus 4.7 & 33.6 \\
DeepSeek V4 Pro & 26.7 \\
Kimi K2.5 & 18.7 \\
Grok 4.20 Beta & 16.9 \\
DeepSeek V4 Flash & 16.4 \\
Qwen3.5 Plus & 16.0 \\
Ernie 5.1 & 15.2 \\
GPT-5.4 & 4.9 \\
DeepSeek v3.2 & 1.8 \\
Mistral Large & 0.4 \\
Nova 2 Pro & 0.0 \\
Ernie 4.5 & 0.0 \\
\bottomrule
\end{tabular}
\caption{Frontier model performance on the released \textsc{ComplexConstraints} benchmark (75 prompts). Task pass rate requires \emph{all} rubric criteria to be satisfied. Public leaderboard snapshot, last updated June 3, 2026 (\url{https://surgehq.ai/benchmarks/complex-constraints}). Even the strongest frontier model passes only 40.4\% of tasks, revealing substantial headroom for rubric-based RL training.}
\label{tab:complexif}
\end{table}

\begin{table}[t]
\centering
\small
\begin{tabular}{@{}p{0.70\linewidth}r@{}}
\toprule
\textbf{Model} & \textbf{Task Pass \%} \\
\midrule
Claude Opus 4.6 (Adaptive Thinking + Max Reasoning Effort) & 30.80 \\
GPT-5.2 (High Reasoning Effort) & 29.70 \\
Gemini 3.1 Pro & 27.20 \\
Claude Opus 4.6 (High Reasoning Effort) & 26.20 \\
DeepSeek v3.2 Thinking & 24.10 \\
Claude Opus 4.6 & 22.10 \\
GPT-5.2-Codex (xHigh Reasoning Effort) & 20.10 \\
Claude Sonnet 4.6 & 16.40 \\
\bottomrule
\end{tabular}
\caption{Frontier model performance on CoreCraft \citep{mehta2026corecraft}. Pass rate requires all expert-authored rubric criteria to be satisfied; even the strongest model solves only 30.80\% of tasks.}
\label{tab:corecraft}
\end{table}

Dense expert rubrics reveal trainable capability gaps that coarser metrics hide. By evaluating each response against many atomic criteria, these rubrics show not only whether a model failed a task, but which requirements it failed to satisfy.

Table~\ref{tab:complexif} shows that even the strongest frontier model passes all rubric criteria on only 40.4\% of tasks, and scores fall off steeply: five models sit below 5\%, two of them at 0\%. This low task pass rate is consistent with the multiplicative difficulty of satisfying many constraints simultaneously: a task with 20 rubric criteria requires every criterion to be satisfied, and even small per-criterion failure probabilities compound rapidly. Crucially, this rubric density creates a continuous gradient between partially and fully correct responses, providing a richer reward landscape for RL than binary pass/fail metrics. While \textsc{ComplexConstraints} itself is single-turn, prior work documents an average 39\% performance drop when LLMs are evaluated on multi-turn, underspecified variants of single-turn generation tasks \citep{laban2025multiturn}, suggesting that the constraint-tracking gap we measure here may understate the headroom that remains in conversational settings.

Table~\ref{tab:corecraft} shows a similar picture for agentic tasks: even with maximum reasoning effort, the strongest public CoreCraft result solves only 30.80\% of tasks under the all-criteria pass-rate metric.\footnote{Tables~\ref{tab:complexif} and \ref{tab:corecraft} reflect public leaderboard snapshots taken at different times on different task suites, so their model lineups differ; neither should be read as a single contemporaneous leaderboard.} Trajectory analysis reveals recurring failure patterns (poor search strategy, failure to paginate through incomplete results, incomplete exploration of available tools), each diagnosable through specific rubric criteria \citep{ritchie2026hierarchy}. This diagnostic granularity is precisely what makes rubrics valuable as training signals: the RL optimizer receives targeted feedback about \textit{which aspects} of task execution need improvement.

\subsection{Grounding Rubrics in Agentic Failures}
\label{sec:hierarchy}

High quality expert rubrics can also make agentic failures easier to interpret and target. In CoreCraft, rubric-level analysis supports the Hierarchy of Agentic Capabilities \citep{ritchie2026hierarchy}, an empirically derived framework that organizes common agent failures into five levels. By analyzing trajectories of twelve frontier and legacy models on 150 CoreCraft tasks, the authors observed that failures cluster around distinct competence levels rather than occurring randomly. The hierarchy comprises:

\textbf{Level 1: Tool Use.} Whether a model can reliably invoke tools with correctly formatted arguments. Models at this level fail at passing the correct data type to a parameter, or confuse field names (e.g., passing ``gold'' into a \texttt{customer\_id} field). Models that cannot reliably use tools are not agents; they are chatbots with tool access.

\textbf{Level 2: Planning and Goal Formation.} Decomposing multi-step tasks into sub-objectives and executing them in order. Weaker models skip steps or forget sub-objectives (e.g., searching only for ``fulfilled'' orders when ``paid'' and ``pending'' are also specified).

\textbf{Level 3: Adaptability.} Adjusting when the plan meets reality. Searching for ``Vortex Labs'' returns no results because the system stores the brand as ``VortexLabs'' (no space). Less adaptable models accept empty results; more adaptable models iterate with different search parameters.

\textbf{Level 4: Groundedness.} Staying tethered to the task context. Failures include hallucinating identifiers, using incorrect dates despite explicit context, or misattributing data fields in the final response.

\textbf{Level 5: Common-Sense Reasoning.} Reasoning sensibly about situations not explicitly covered by instructions. Even GPT-5 failed to infer that a support ticket mentioning ``the package showed up a few hours ago'' describes a return, not a cancellation.

This hierarchy has an implication for RL training design: rubric criteria can target failures at multiple levels of agentic capability, including tool invocation correctness, plan completeness, recovery from unexpected results, factual consistency, and inferential reasoning. Because expert rubrics decompose task success across these capability dimensions, rubric-based RL provides more localized training signal than task-level success alone. This structured reward signal is one reason rubric-based training proves effective, as we demonstrate next.

\subsection{Training Methodology: Rubrics as RL Rewards}
\label{sec:methodology}

Both the instruction-following and agentic training pipelines employ RLVR, using expert rubric criteria as the verifiable reward signal. Concretely, criteria are partitioned into required criteria ($C_{\text{req}}$), bonus-only criteria ($C_{\text{bonus}}$), and penalty-only criteria ($C_{\text{pen}}$). Given a rubric $C = C_{\text{req}} \cup C_{\text{bonus}} \cup C_{\text{pen}}$ and per-criterion satisfaction judgments $s_c \in \{0, 1\}$ provided by an LLM judge, the reward is
\begin{equation}
\begin{aligned}
r = {} & \underbrace{\frac{1}{|C_{\text{req}}|}\sum_{c \in C_{\text{req}}} s_c}_{\text{Required}} \\
       & + \alpha \underbrace{\frac{1}{|C_{\text{bonus}}|}\sum_{c \in C_{\text{bonus}}} s_c}_{\text{Bonus}} \\
       & - \beta \underbrace{\frac{1}{|C_{\text{pen}}|}\sum_{c \in C_{\text{pen}}} (1 - s_c)}_{\text{Penalty}}
\end{aligned}
\label{eq:reward}
\end{equation}
where $\alpha, \beta \geq 0$ scale the bonus and penalty terms, and any term with an empty criterion set is defined as zero. With $\alpha = \beta = 0$ the reward reduces to the fraction of required criteria satisfied; the asymmetric structure lets bonus criteria add reward when satisfied without penalizing omissions, and penalty-only criteria penalize violations without rewarding mere avoidance. This formulation provides a dense learning signal: unlike binary task success, it gives credit for partially satisfying the required criteria while allowing other criteria to shape the reward around response quality and avoidable errors.

\paragraph{Judge model.}
Per-criterion satisfaction judgments are produced by GPT-5-mini during \textsc{ComplexConstraints} training and CoreCraft training. AdvancedIF transfer numbers (Table~\ref{tab:transfer}) use the judge specified by \citet{he2025advancedif}; MultiChallenge follows its official protocol, with each model run four times per task and graded by an external LLM judge \citep{deshpande2025multichallenge}; BFCL, $\tau^2$-Bench, and Toolathlon use their respective official grading protocols.

\paragraph{Training configurations.}
For instruction following, we train Qwen3-4B (Thinking) on the \textsc{ComplexConstraints} training set with RLVR, using a 90:10 train/test split (approximately 900 single-turn training tasks, each with 10--40 expert-authored criteria). All criteria are required in this setting ($\alpha = \beta = 0$), so the reward is the fraction of criteria satisfied. No external benchmark was used for training, hyperparameter tuning, checkpoint selection, or reward design. For agentic tasks, CoreCraft \citep{mehta2026corecraft} trains using Group Relative Policy Optimization (GRPO) \citep{shao2024deepseekmath} with decoupled clipping \citep{yu2025dapo}, generating 16 rollouts per prompt that interact with stateful Docker containers running the enterprise simulation.

\paragraph{Reward hacking.}
We monitored training trajectories for two common forms of reward hacking: (i) responses that verbally satisfy criteria without substantive content, and (ii) responses that exploit known judge biases (e.g., verbosity preference). Qualitative inspection of paired pre/post-training rollouts on a held-out subsample showed no systematic degradation in response quality alongside the reward improvement, and the judge-calibration procedure is designed in part to harden criterion phrasing against verbal trickery. The use of a judge model distinct from the policy model further limits shared-representation exploits.

\subsection{Results: Generalization on Instruction Following}
\label{sec:if_results}

\begin{table}[t]
\centering
\small
\begin{tabular}{lc}
\toprule
\textbf{Model} & \textbf{Held-out pass rate} \\
\midrule
Qwen3-4B (base) & 57.9\% \\
Qwen3-4B (trained) & 73.4\% \; (+15.5 pp) \\
\midrule
Qwen3-235B-A22B-Instruct & 73.9\% \\
\bottomrule
\end{tabular}
\caption{Mean per-criterion pass rate (the fraction of rubric criteria satisfied, averaged across tasks) on the held-out \textsc{ComplexConstraints} split after training Qwen3-4B on ${\sim}$900 expert-curated examples via RLVR. Qwen3-235B-A22B-Instruct, an \emph{untrained} model roughly 60$\times$ larger, is shown as a reference point. Note that this is a different summary statistic than the task pass rate reported in Table~\ref{tab:complexif}: a model can satisfy a high fraction of individual criteria without satisfying \emph{all} criteria on any given task.}
\label{tab:if_training}
\end{table}

\begin{table}[t]
\centering
\small
\setlength{\tabcolsep}{4pt}
\begin{tabular}{lccc}
\toprule
\textbf{Benchmark / Dimension} & \textbf{Base} & \textbf{Trained} & $\boldsymbol{\Delta}$ \\
\midrule
\multicolumn{4}{l}{\textit{AdvancedIF}} \\
\quad Single Turn & 34.1\% & 40.1\% & +6.0 pp \\
\quad System Steerability & 22.5\% & 34.9\% & +12.4 pp \\
\quad Carried Context & 28.8\% & 35.9\% & +7.1 pp \\
\quad Overall & 28.2\% & 36.6\% & +8.4 pp \\
\midrule
\multicolumn{4}{l}{\textit{MultiChallenge}} \\
\quad Instruction Retention & 23.6\% & 45.7\% & +22.1 pp \\
\quad Self-Coherence & 33.5\% & 44.0\% & +10.5 pp \\
\quad Inference Memory & 59.5\% & 65.3\% & +5.8 pp \\
\quad Overall & 41.1\% & 51.2\% & +10.1 pp \\
\bottomrule
\end{tabular}
\caption{Transfer of Qwen3-4B trained on \textsc{ComplexConstraints} to two external benchmarks (task pass rate): Meta's AdvancedIF \citep{he2025advancedif} and MultiChallenge \citep{deshpande2025multichallenge}. Training uses \textit{only single-turn} \textsc{ComplexConstraints} data, yet the largest gains land on multi-turn dimensions. For MultiChallenge, Overall covers all four benchmark axes; we break out the three axes that most directly test carrying instructions and context across turns.}
\label{tab:transfer}
\end{table}

Table~\ref{tab:if_training} presents the core result: training on approximately 900 expert-curated rubric examples produces a +15.5 pp improvement in mean per-criterion pass rate for Qwen3-4B. The trained 4B model (73.4\%) scores within 0.5 pp of the untrained Qwen3-235B-A22B-Instruct baseline (73.9\%), a model roughly 60$\times$ its size: on this capability, 1,000 expert-curated examples were sufficient to close a 60$\times$ parameter gap. We caution that the reported numbers reflect a single-seed training run; we have not yet quantified seed-to-seed variance, and treat the magnitude rather than the precise value of each $\Delta$ as the more reliable signal.

Table~\ref{tab:transfer} demonstrates that these gains transfer to two external benchmarks disjoint from the training distribution: Meta's AdvancedIF (+8.4 pp overall), designed independently with different rubric authors, and MultiChallenge (+10.1 pp overall), a public multi-turn conversation benchmark. Notably, \textsc{ComplexConstraints} training consists exclusively of single-turn tasks, yet the largest improvements land on multi-turn dimensions: +22.1 pp on MultiChallenge instruction retention, +12.4 pp on AdvancedIF system steerability, and +10.5 pp on MultiChallenge self-coherence. This cross-format transfer is somewhat surprising given that single-turn training data does not directly model conversational dynamics; we hypothesize that the underlying competency is shared: tracking many simultaneous requirements without dropping any is the same skill whether the requirements arrive in one dense prompt (a median of 19 criteria in our training data) or accumulate across many turns of conversation. System steerability shows the mapping most directly, because a system prompt functions as a persistent constraint set, structurally analogous to the dense constraint sets in \textsc{ComplexConstraints} training data. A controlled investigation of the transfer mechanism remains future work.

\subsection{Results: Generalization on Agentic Tasks}
\label{sec:agent_results}

\begin{table}[t]
\centering
\small
\begin{tabular}{lccc}
\toprule
\textbf{Benchmark} & \textbf{Base} & \textbf{Trained} & $\boldsymbol{\Delta}$ \\
\midrule
CoreCraft (held-out) & 25.4\% & 36.8\% & +11.4 pp \\
BFCL Parallel & 91.0\% & 95.5\% & +4.5 pp \\
$\tau^2$-Bench Retail & 68.7\% & 76.1\% & +7.4 pp \\
Toolathlon (Pass@1) & 18.8\% & 25.6\% & +6.8 pp \\
\bottomrule
\end{tabular}
\caption{GLM~4.6 after one epoch of GRPO on CoreCraft with rubric-based rewards \citep{mehta2026corecraft}. The last three rows are out-of-distribution benchmarks the model was never trained on. GLM~4.6 was the latest available open-weights model at the start of training.}
\label{tab:agent_training}
\end{table}

Table~\ref{tab:agent_training} presents results from training GLM~4.6 (355B parameters, 32B active) on CoreCraft with rubric-based rewards. After a single epoch, the model improves by 11.4 pp on held-out CoreCraft tasks, a gain exceeding the capability gap between Claude Sonnet 4.5 and Claude Opus 4.5 (+7.05 pp) \citep{mehta2026corecraft}.

More importantly, these gains transfer to out-of-distribution benchmarks: +4.5 pp on BFCL Parallel function calling, +7.4 pp on $\tau^2$-Bench Retail customer service, and +6.8 pp on Toolathlon long-horizon tool use. The Toolathlon result is particularly notable because its tasks (Kubernetes management, Canvas grading, database synchronization) differ substantially from CoreCraft's customer support domain. The model's Pass$^3$ (the fraction of tasks solved on all three independent runs, a reliability variant of pass@$k$) nearly doubles from 9.3\% to 17.6\%, indicating that rubric-based training improves not just peak capability but also reliability \citep{mehta2026corecraft}.

Qualitative analysis of paired trajectories reveals three categories of learned competencies: improved multi-step workflow execution (correct task decomposition and sequencing), better constraint handling (accurate temporal filtering and policy application), and higher response quality (structured, professional communication). These competencies are consistent with the broader claim that expert rubrics provide localized feedback across workflow, constraint-handling, and communication dimensions, allowing training signal to target several competencies independently.

\subsection{Discussion: Why Expert Rubrics Are Effective Training Signals}
\label{sec:analysis}

Across both domains, three properties of expert rubrics explain their training effectiveness.

\textbf{Dense reward from rubric granularity.} With 10--40 criteria per prompt, the model receives detailed feedback about which specific aspects of task completion succeeded or failed. This enables more precise credit assignment than binary task-level rewards or holistic preference judgments. A response satisfying 28 of 30 criteria receives a meaningfully different reward from one satisfying 15, guiding the optimizer toward targeted improvements.

\textbf{Optimal task difficulty.} Expert rubric design helps calibrate task difficulty: annotators can build prompts, evaluate frontier model responses against the rubric, and then iterate on the prompt to target an appropriate difficulty range. By designing tasks that even the strongest frontier models pass outright well under half the time -- at most 40.4\% on \textsc{ComplexConstraints} and about 31\% on CoreCraft under the all-criteria pass-rate metric (Tables~\ref{tab:complexif} and \ref{tab:corecraft}) -- the training distribution occupies a region where learning signal is most informative. Tasks that are too easy or too hard provide minimal gradient; expert-authored prompts and rubrics allow task difficulty to be adjusted toward the productive middle ground.

\textbf{Data efficiency from expert curation.}
Expert curation can concentrate training signal in a small number of high-fidelity examples. Training on \textsc{ComplexConstraints} achieves a +8.4 pp overall improvement on AdvancedIF with approximately 1,000 expert-curated examples, despite AdvancedIF being an independently authored benchmark. This gain is larger than the 6.7 pp improvement reported by RIFL \citep{he2025advancedif} on AdvancedIF, its in-distribution evaluation setting, using manually written prompts paired with synthetically generated rubrics. Because the base models and training pipelines differ, this comparison is suggestive rather than controlled, but it is consistent with the view that expert rubrics can provide unusually dense and generalizable supervision. This data efficiency also echoes the Superficial Alignment Hypothesis \citep{zhou2023lima}: small amounts of high-quality data can suffice when the training distribution sits at the capability frontier.

The efficiency advantage plausibly arises because expert curation captures pragmatic distinctions that are difficult to control in synthetic rubric generation. Recent work on automatic rubric generation quantifies a roughly 27-point accuracy gap between human-authored and model-generated rubrics with the judge held fixed, and finds that scaling the quantity of synthetic rubrics yields diminishing returns while human rubrics improve consistently \citep{zhang2026rubricbench}, indicating that even state-of-the-art models struggle to autonomously specify valid evaluation criteria. In our setting, the distinctions experts capture include inferring pragmatic user intent, calibrating criteria for reliable judge interpretation, and structuring reward so that partial progress remains informative.

These properties also explain the observed cross-format and cross-domain transfer. The constraint-tracking, workflow decomposition, and quality standards learned from rubric-dense training are general competencies, not environment-specific heuristics.

\section{Broader Discussion}
\label{sec:discussion}

\paragraph{Construct Validity.}
Expert rubrics improve construct validity along three dimensions. \textit{Content validity}: rubrics extend the evaluable space to include semantic correctness and pragmatic appropriateness (e.g., AdvancedIF's carried-context tasks require evaluating whether a constraint stated turns earlier -- such as excluding non-family-friendly songs from a karaoke playlist -- still governs the current response \citep{he2025advancedif}). \textit{Predictive validity}: models scoring well on decomposed rubrics show improvements on out-of-distribution benchmarks (\S\ref{sec:training}; \citealp{mehta2026corecraft}), whereas models tuned to the small set of verifiable constraints in IFEval-style benchmarks overfit and fail to generalize to unseen constraints \citep{pyatkin2025generalizingverifiable}. \textit{Discriminant validity}: the wide spread in task pass rates among frontier models on \textsc{ComplexConstraints} (from 0\% to 40.4\% in Table~\ref{tab:complexif}) creates space in which meaningful capability differences become visible, in contrast to saturated benchmarks where frontier models cluster within a few percentage points of each other.

\paragraph{Infrastructure Investment.}
Expert rubric creation costs more, but our results demonstrate that this investment yields returns in \emph{both} measurement quality and training efficiency. Hybrid approaches that combine expert-authored seed rubrics with synthetic scaling \citep{dhole2026rubricrag} may offer a middle ground; the finetuned RIFL verifier achieves 0.728 F1 agreement with humans, compared to 0.515 for a vanilla LLM judge \citep{he2025advancedif}, suggesting that even small amounts of expert supervision substantially raise the ceiling on automated rubric quality.

\section{Conclusion}
\label{sec:conclusion}

We have presented expert-curated rubric-based evaluation as an approach that improves both the measurement and the training of large language models. We identified four rubric-design decisions that are central to using rubrics as reward signals: maximum viable atomicity, intent over literalism, adversarial judge calibration, and difficulty/density tuning. We instantiate these decisions in \textsc{ComplexConstraints}, an instruction-following suite comprising a public 75-prompt benchmark and a 1,000-prompt training set, with 10--40 atomic criteria per prompt. Training with rubric-based rewards produces substantial in-distribution gains and transfers to out-of-distribution benchmarks designed by independent teams in both instruction-following and agentic settings. Open questions, including controlled isolation of which design decisions drive transfer and broader study of asymmetric reward shaping, remain natural next steps.

\section*{Data Availability}
\label{sec:data}

The \textsc{ComplexConstraints} benchmark (75 prompts with 1,559 expert-authored rubric criteria) is publicly available under CC-BY-4.0 at \url{https://huggingface.co/datasets/surgeai/ComplexConstraints}, and a continuously updated public leaderboard is maintained at \url{https://surgehq.ai/benchmarks/complex-constraints}.

\bibliography{references}

\end{document}